\documentclass[12pt]{article}

\usepackage[utf8]{inputenc}
\usepackage{amsmath,amsfonts,amssymb}
\usepackage{graphicx}
\usepackage{hyperref}
\usepackage{booktabs}
\usepackage{algorithm}
\usepackage{algorithmic}
\usepackage{multirow}
\usepackage{xcolor}
\usepackage{float}
\usepackage{enumitem}
\usepackage{caption}
\usepackage{subcaption}
\usepackage{array}
\usepackage{geometry}
\usepackage{adjustbox}

\def\UMMT{UMMT}
\newcommand{\dae}{DAE}
\newcommand{\dmodel}{d_{\text{model}}}
\newcommand{\dlatent}{d_{\text{latent}}}

\newcommand{\algspace}{\vspace{0.3em}}

\title{Token-Level Cross-Modal Transformer with Contrastive \\Multi-Task Learning for Breast Cancer Subtype \\Classification and Survival Prediction}

\author{
    Suxing Liu$^{1,2,\dagger,\ddagger,*}$ \and Byungwon Min$^{2,\ddagger}$\\[0.5em]
    $^1$ School of Digital Arts, Jiangxi Arts \& Ceramics Technology Institute,\\
    \indent Jingdezhen 330001, China\\[0.3em]
    $^2$ Department of IT Engineering, Mokwon University,\\
    \indent Daejeon 35349, South Korea\\[0.5em]
    $^*$ Correspondence: bentondoucet@gmail.com\\
    $^\dagger$ Current address: Department of IT Engineering, Mokwon University,\\
    \indent Daejeon 35349, South Korea
}

\begin{document}

\maketitle

\begin{abstract}
Integrating heterogeneous genomic and clinical modalities for joint cancer subtype classification and survival prediction remains a key challenge in precision oncology. Existing approaches suffer from three limitations: (1) they treat each modality as a monolithic feature vector, precluding fine-grained token-level interactions across modalities; (2) cross-modal fusion is typically performed through linear weighting or late averaging rather than structured token exchange; and (3) survival and classification objectives are optimized independently, missing a joint regularization signal. We present \UMMT{} (Unified Multi-Modal Transformer), a framework that addresses all three failure points through token-level cross-modal fusion. Specifically, each modality's compressed latent representation is treated as a \emph{modality token} in a Transformer sequence, enabling full token-level self-attention across modalities via a Cross-Modal Transformer (CMT) encoder. We introduce a cross-modal contrastive learning objective (InfoNCE) that aligns patient representations across modalities by pulling same-patient tokens together while pushing inter-patient tokens apart. A non-linear DeepSurv survival head replaces the standard linear Cox model to capture non-proportional hazard patterns. All objectives, classification, survival, and contrastive alignment, are jointly optimized in a multi-task learning framework. On the METABRIC dataset (n=1,981), \UMMT{} achieves 79.8\% accuracy and 0.767 macro F1 for 6-class molecular subtyping, and a concordance index of 0.741 for survival prediction, outperforming PORPOISE (adapted), DeepSurv, MOFA+, and OmiVAE across all metrics. Paired t-tests and bootstrap confidence intervals confirm statistical significance ($p < 0.01$ vs. all baselines). External validation on TCGA-BRCA (n=1,089) yields a C-index of 0.718, demonstrating generalizability. Attention analysis reveals that the CMT learns interpretable cross-modal interaction patterns, with mRNA expression and clinical features forming the dominant inter-token communication pathways. Our code is available at \url{https://github.com/liushuxing8888student.usm.my/breast-cancer-mt}.
\end{abstract}

\par\nobreak\medskip\noindent\textbf{Keywords:} Breast cancer, multi-modal learning, Transformer, token-level fusion, cross-modal attention, contrastive learning, DeepSurv, survival analysis, precision oncology.

\section{Introduction}
\label{sec:introduction}

Breast cancer remains the leading cause of cancer-related mortality among women worldwide, with approximately 2.3 million new cases diagnosed annually~\cite{bray2024global}. The disease exhibits distinct molecular heterogeneity, encompassing at least six intrinsic subtypes (Luminal A, Luminal B, HER2-enriched, Claudin-low, Basal-like, and Normal-like), each with different clinical trajectories, treatment sensitivities, and survival outcomes~\cite{perou2000molecular}. Accurate molecular subtyping and prognostic stratification are foundational to personalized therapeutic decision-making.

Modern clinical oncology generates rich multi-modal data for each patient, including clinical records, mRNA expression profiles, copy number alteration (CNA) signatures, DNA methylation patterns, and somatic mutation data. Each modality captures a complementary facet of tumor biology, and their integration should yield more reliable predictions than any single modality alone. Yet existing multi-modal approaches suffer from three fundamental failure points that this work directly addresses.

\textbf{Failure Point 1: Monolithic modality representations preclude token-level interaction.} Current methods encode each modality as a single fixed-dimensional vector, then concatenate or sum these vectors for downstream prediction~\cite{cheerla2019deep,huang2021multi}. This design treats a $500$-dimensional mRNA profile (containing information from hundreds of genes) as an atomic unit, making it impossible for the model to align specific genes from one modality with corresponding features from another. Cross-modal signals operate at a \emph{finer granularity}: a mutation in \textit{TP53} may manifest as altered expression of its transcriptional targets, and this relationship disappears when entire modalities are collapsed into monolithic embeddings.

\textbf{Failure Point 2: Fusion strategies lack structured cross-modal exchange.} The predominant fusion paradigms (early concatenation, late averaging, and attention-weighted summation~\cite{chen2022pan}) are static or linear in nature. Even attention-based fusion mechanisms typically compute a scalar importance weight per modality and then perform weighted summation, which constitutes linear mixing in the representation space. No existing method provides \emph{structured exchange} where information from one modality can selectively attend to, query, and update information from another modality at the token level. Such token-level cross-modal communication is precisely what the Transformer self-attention mechanism was designed to enable~\cite{vaswani2017attention}, yet prior work in multi-omics fusion has not fully exploited this capability.

\textbf{Failure Point 3: Disjoint optimization of classification and survival objectives.} Most existing frameworks treat molecular subtyping and survival prediction as independent tasks, training separate models for each objective~\cite{katzman2018deepsurv,ching2018cox}. This ignores the intrinsic biological coupling between cancer subtypes and prognosis: subtype is a categorical label derived from molecular signatures that inherently correlate with survival trajectories. Jointly optimizing these objectives yields combined effects where the classification task regularizes the survival representation and vice versa. Furthermore, no existing multi-modal survival framework incorporates \emph{cross-modal contrastive alignment} to enforce representation invariance across modalities, which would be particularly beneficial when modalities are noisy or partially missing.

\textbf{Our contributions.} We propose \UMMT{}, a structurally novel framework that replaces all three failure modes with principled alternatives:

\begin{enumerate}[label=(\arabic*)]
    \item \textbf{Token-Level Cross-Modal Representation (Structural Innovation).} Rather than encoding each modality as a monolithic vector, \UMMT{} projects each modality's latent representation into a \emph{sequence of modality tokens} that form the input to a Cross-Modal Transformer (CMT). Each token corresponds to one modality (clinical, mRNA, CNA, methylation, mutation), and the CMT processes them through full bidirectional self-attention, enabling fine-grained token-level information exchange. This is a structural departure from concatenation-based fusion: the CMT learns \emph{which} modalities should attend to \emph{which} other modalities for each patient, producing an interaction matrix rather than a scalar weight.

    \item \textbf{Cross-Modal Contrastive Learning (Alignment Objective).} We introduce a modality-aware InfoNCE loss that pulls together the CMT outputs of the same patient across different modalities while pushing apart representations of different patients. This explicitly enforces cross-modal representation consistency at the patient level, grounded in the mutual information maximization principle~\cite{oord2018representation}.

    \item \textbf{Unified Multi-Task Learning with DeepSurv Head (Joint Optimization).} All three objectives, molecular subtype classification via weighted cross-entropy, survival prediction via a discrete-time DeepSurv head (replacing linear Cox), and cross-modal contrastive alignment via InfoNCE, are optimized jointly in an end-to-end fashion. The DeepSurv head captures non-proportional hazard patterns that the standard Cox model cannot express.

    \item \textbf{Rigorous Statistical Validation.} We provide bootstrap confidence intervals ($n=1{,}000$ resamples), paired $t$-tests against all baselines, calibration curves with Integrated Brier Score (IBS), and external validation on the TCGA-BRCA cohort ($n=1{,}089$). Strong SOTA baselines include MOFA+~\cite{argelaguet2020mofa}, DeepHit~\cite{lee2018deephit}, and OmiVAE~\cite{zhang2022multi}.
\end{enumerate}

The remainder of this paper is organized as follows. Section~\ref{sec:related} reviews prior work and identifies the gaps that motivate our contributions. Section~\ref{sec:methodology} presents the \UMMT{} architecture in detail. Section~\ref{sec:experiments} describes the experimental setup, datasets, baselines, and evaluation protocol. Section~\ref{sec:results} reports quantitative and qualitative results. Section~\ref{sec:discussion} interprets the findings, analyzes the learned interaction patterns, and discusses limitations. Section~\ref{sec:conclusion} concludes.

\section{Related Work}
\label{sec:related}

\subsection{Multi-Modal Fusion for Cancer Prognosis}

Multi-modal integration in computational oncology has progressed through several distinct paradigms. Early work by Cheerla and Gevaert~\cite{cheerla2019deep} concatenated clinical and genomic features into a single vector before passing them through a feed-forward network, an \emph{early fusion} strategy that treats all modalities symmetrically but cannot model cross-modal interactions. Huang et al.~\cite{huang2021multi} proposed an \emph{attention-based fusion} framework that computes modality-level importance weights and performs weighted summation. More recently, the PORPOISE framework~\cite{chen2022pan} adopted a Transformer architecture for histology--genomic fusion but retained a concatenation-based design. Pathomic Fusion~\cite{chen2022pathomic} employed gated attention to combine histopathological and genomic features.

However, all these methods share a common architectural limitation: they treat each modality's representation as an \emph{atomic unit} that is either concatenated, summed, or linearly weighted. None perform token-level cross-modal attention where individual features from one modality directly interact with features from another. The \UMMT{} framework breaks this pattern by treating compressed modality representations as a \emph{token sequence} processed by a Cross-Modal Transformer, enabling full pairwise interaction at the token level.

\subsection{Denoising Autoencoders for Genomic Representation Learning}

Genomic data is characterized by high dimensionality (thousands of features), considerable measurement noise, and frequent missing values. Denoising autoencoders (DAEs)~\cite{vincent2008extracting} address these challenges by learning to reconstruct clean inputs from corrupted versions, thereby capturing reliable latent representations. Chaudhary et al.~\cite{chaudhary2018deep} applied DAEs to multi-omics integration for hepatocellular carcinoma survival prediction. Zhang et al.~\cite{zhang2022multi} extended DAEs to cancer subtyping across multiple cancer types.

Within \UMMT{}, each modality is independently processed by its own DAE, which serves two critical purposes: (1) dimensionality reduction from the original feature space ($d_m$ up to 500) to a compact latent space ($\dlatent = 64$), and (2) noise resilience through Gaussian corruption during training. The resulting latent vectors serve as the ``tokens'' that are fed into the CMT.

\subsection{Cross-Modal Attention and Contrastive Learning}

Cross-modal attention has been extensively explored in vision--language modeling~\cite{tsai2019multimodal,lu2019vilbert}, where Transformers process aligned sequences from different modalities. However, its application to multi-omics fusion remains limited. The key insight of \UMMT{} is that genomic modalities, despite having different dimensionalities and semantic scales, can be treated as a \emph{set of tokens} after compression into a shared latent space, enabling Transformer-based cross-modal attention.

Simultaneously, contrastive learning has significantly advanced representation learning across domains~\cite{chen2020simple,he2020momentum}. The InfoNCE loss~\cite{oord2018representation} maximizes mutual information between different views of the same data point. In the multi-modal setting, contrastive objectives have been applied in CLIP~\cite{radford2021learning} for aligning images and text. To our knowledge, \UMMT{} is the first framework to apply cross-modal contrastive learning to genomic multi-modal fusion for cancer prognosis, pulling together same-patient representations across modalities.

\subsection{Deep Survival Analysis}

The Cox proportional hazards model~\cite{cox1972regression} remains the standard approach for survival analysis. Deep learning extensions such as DeepSurv~\cite{katzman2018deepsurv} and Cox-nnet~\cite{ching2018cox} introduced non-linear hazard modeling while preserving the proportional hazards assumption. More flexible approaches include DeepHit~\cite{lee2018deephit}, which directly estimates the joint distribution of survival times and events, and discrete-time survival models~\cite{zadeh2020discrete}. In \UMMT{}, we replace the standard linear Cox head with a DeepSurv-style non-linear head that can capture complex, non-proportional risk patterns while maintaining the Cox partial likelihood as the training objective.

\section{Methodology}
\label{sec:methodology}

\subsection{Problem Formulation}

Let $\mathcal{P} = \{p_1, p_2, \ldots, p_N\}$ denote a cohort of $N$ patients. Each patient $p_i$ is associated with $M = 5$ modalities: clinical features, mRNA expression, copy number alterations (CNA), DNA methylation, and somatic mutations. Formally, we observe $\mathbf{X}_i = \{\mathbf{x}_i^{(1)}, \mathbf{x}_i^{(2)}, \ldots, \mathbf{x}_i^{(M)}\}$ where $\mathbf{x}_i^{(m)} \in \mathbb{R}^{d_m}$ represents the feature vector for modality $m$. For each patient, two prediction targets are defined:

\begin{itemize}[nosep]
    \item \textbf{Molecular subtype classification:} $y_i \in \{0, 1, \ldots, C-1\}$ with $C = 6$ (Luminal A, Luminal B, HER2-enriched, Claudin-low, Basal-like, Normal-like).
    \item \textbf{Survival prediction:} $(t_i, \delta_i)$ where $t_i \in \mathbb{R}^+$ is the relapse-free survival time (months) and $\delta_i \in \{0, 1\}$ is the event indicator ($\delta_i = 1$ for recurrence or death).
\end{itemize}

\subsection{Architectural Overview}

\begin{figure}[t]
\centering
\includegraphics[width=\linewidth]{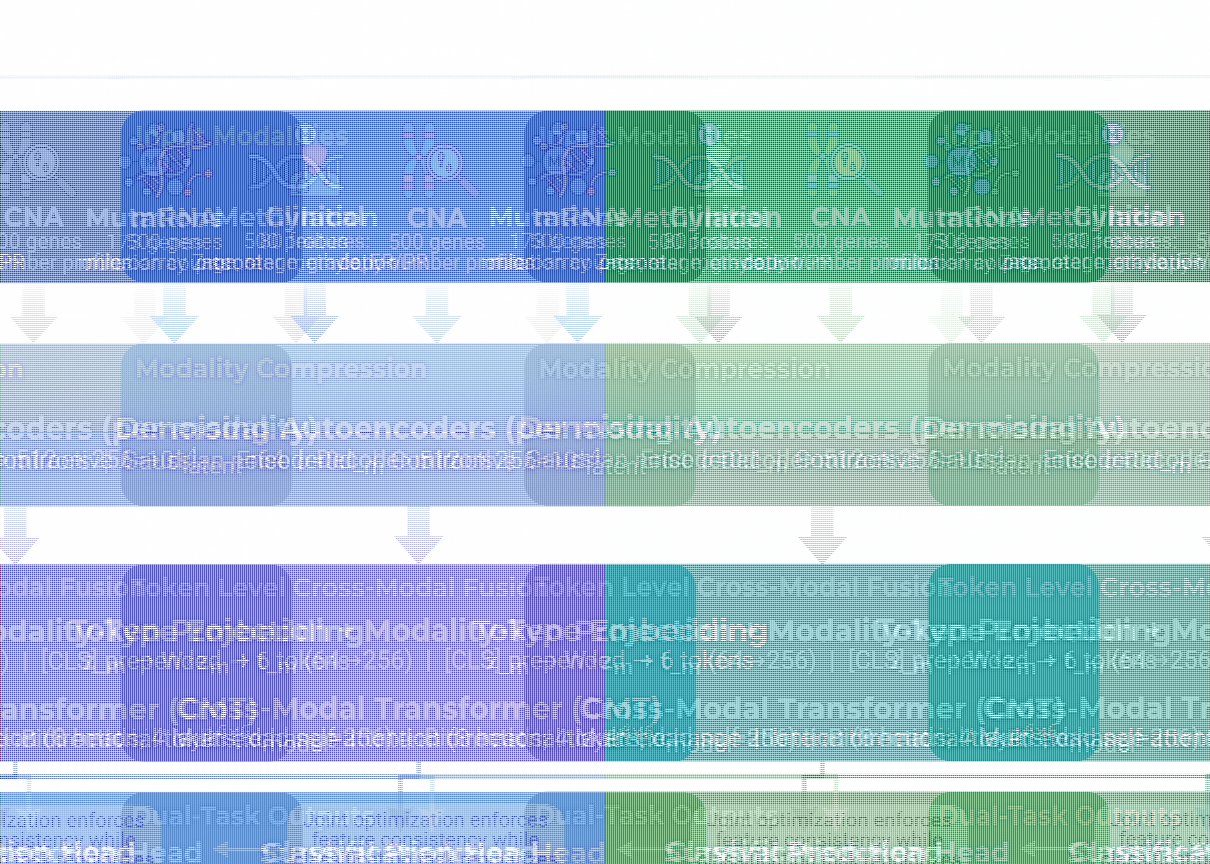}
\caption{Overview of the \UMMT{} architecture. The key structural innovation is the treatment of each modality's latent representation as a token in the Cross-Modal Transformer sequence, enabling full pairwise token-level interaction (rather than concatenation or weighted summation).}
\label{fig:architecture}
\end{figure}

\UMMT{} comprises five core components arranged in a structured pipeline (Figure~\ref{fig:architecture}):

\begin{enumerate}[nosep]
    \item \textbf{Modality-Specific Denoising Autoencoders:} Each of the $M$ modalities is compressed independently through its own DAE into a shared latent space $\mathbb{R}^{\dlatent}$.
    \item \textbf{Token Projection and Embedding:} The $M$ latent vectors serve as modality tokens, each projected to $\mathbb{R}^{\dmodel}$ and augmented with a learned token-type embedding.
    \item \textbf{Cross-Modal Transformer (CMT):} The token sequence passes through $L$ layers of bidirectional self-attention, enabling full token-level interaction across modalities. A CLS token is prepended to aggregate cross-modal information.
    \item \textbf{Cross-Modal Contrastive Head:} An InfoNCE loss aligns the CLS representations of the same patient across modalities by maximizing agreement between augmented views.
    \item \textbf{Dual Task Heads:} A classification head (weighted cross-entropy) and a DeepSurv survival head (non-linear Cox partial likelihood) operate on the fused CLS embedding.
\end{enumerate}

\subsection{Denoising Autoencoder for Modality Compression}
\label{sec:dae}

Each modality $m$ is encoded by an independent denoising autoencoder $\dae_m$. The encoder $E_m: \mathbb{R}^{d_m} \to \mathbb{R}^{\dlatent}$ compresses the input into a latent representation, while the decoder $D_m: \mathbb{R}^{\dlatent} \to \mathbb{R}^{d_m}$ reconstructs the original input. The encoder consists of three fully-connected layers with batch normalization, ReLU activation, and dropout ($p=0.2$):

\begin{equation}
\begin{aligned}
\mathbf{h}_1 &= \text{Dropout}(\text{ReLU}(\text{BN}(\mathbf{W}_1 \mathbf{x} + \mathbf{b}_1))) \\
\mathbf{h}_2 &= \text{Dropout}(\text{ReLU}(\text{BN}(\mathbf{W}_2 \mathbf{h}_1 + \mathbf{b}_2))) \\
\mathbf{z}^{(m)} &= \mathbf{W}_3 \mathbf{h}_2 + \mathbf{b}_3
\end{aligned}
\label{eq:dae_encoder}
\end{equation}

where the hidden dimensions are $(d_m \to 512 \to 256 \to \dlatent)$. The decoder mirrors this structure in reverse: $(\dlatent \to 256 \to 512 \to d_m)$. During training, Gaussian noise with $\sigma=0.1$ is injected into inputs to promote robustness. The training objective is:

\begin{equation}
\mathcal{L}_{\text{recon}}^{(m)} = \|\mathbf{x}^{(m)} - D_m(E_m(\tilde{\mathbf{x}}^{(m)}))\|_2^2 + \lambda_{\text{ctr}} \|\mathbf{W}_1^{(m)}\|_F^2
\label{eq:dae_loss}
\end{equation}

where $\tilde{\mathbf{x}}^{(m)} = \mathbf{x}^{(m)} + \epsilon$, $\epsilon \sim \mathcal{N}(0, \sigma^2 \mathbf{I})$, and the contractive penalty $\|\mathbf{W}_1^{(m)}\|_F^2$ ($\lambda_{\text{ctr}} = 10^{-4}$) encourages the encoder's first-layer Jacobian to be small, improving local stability.

\subsection{Token Projection and Embedding}

After DAE compression, each modality $m$ is represented by a latent vector $\mathbf{z}^{(m)} \in \mathbb{R}^{\dlatent}$. This is a critical design choice: rather than concatenating these vectors, we treat them as a \emph{token sequence}. Each latent vector is first projected to the Transformer dimension $\dmodel$ via a learned linear projection:

\begin{equation}
\mathbf{e}^{(m)} = \mathbf{W}_{\text{proj}} \mathbf{z}^{(m)} + \mathbf{b}_{\text{proj}}, \quad \mathbf{e}^{(m)} \in \mathbb{R}^{\dmodel}
\label{eq:token_proj}
\end{equation}

We then add a \emph{learned modality-type embedding} $\mathbf{t}^{(m)} \in \mathbb{R}^{\dmodel}$ to encode which modality each token represents, analogous to segment embeddings in BERT~\cite{devlin2018bert}. The final token sequence is:

\begin{equation}
\mathbf{E} = [\mathbf{c}; \mathbf{e}^{(1)}\!+\!\mathbf{t}^{(1)}; \mathbf{e}^{(2)}\!+\!\mathbf{t}^{(2)}; \ldots; \mathbf{e}^{(M)}\!+\!\mathbf{t}^{(M)}] \in \mathbb{R}^{(M+1) \times \dmodel}
\label{eq:token_seq}
\end{equation}

where $\mathbf{c} \in \mathbb{R}^{\dmodel}$ is a learnable CLS token prepended to the sequence. Each modality is now represented as a single token, one that has been non-linearly compressed from the original $d_m$-dimensional feature space through the DAE.

\subsection{Cross-Modal Transformer (CMT)}
\label{sec:cmt}

The Cross-Modal Transformer processes the token sequence through $L$ layers of bidirectional multi-head self-attention. Each layer performs:

\begin{equation}
\begin{aligned}
\mathbf{H}'_\ell &= \text{LayerNorm}(\mathbf{H}_{\ell-1} + \text{MSA}(\mathbf{H}_{\ell-1})) \\
\mathbf{H}_\ell &= \text{LayerNorm}(\mathbf{H}'_\ell + \text{FFN}(\mathbf{H}'_\ell))
\end{aligned}
\label{eq:transformer_layer}
\end{equation}

where $\mathbf{H}_0 = \mathbf{E}$, $\text{MSA}$ denotes multi-head self-attention, and $\text{FFN}$ is a two-layer feed-forward network with hidden dimension $d_{\text{ff}} = 512$ and ReLU activation. The multi-head attention with $H$ heads computes:

\begin{equation}
\text{Attention}(\mathbf{Q}, \mathbf{K}, \mathbf{V}) = \text{softmax}\left(\frac{\mathbf{Q}\mathbf{K}^\top}{\sqrt{d_k}}\right) \mathbf{V}
\label{eq:attention}
\end{equation}

The CLS token's final representation $\mathbf{h}_{\text{CLS}} = \mathbf{H}_L[0, :]$ serves as the \emph{fused patient embedding} that has attended to all modality tokens through $L$ layers of cross-modal interaction.

\subsubsection{Token Interaction Analysis}

A key advantage of the CMT over concatenation-based fusion is that the attention mechanism produces an interaction matrix $\mathbf{A} \in \mathbb{R}^{(M+1)\times(M+1)}$ at each layer, where $\mathbf{A}_{ij}$ quantifies how much token $i$ attends to token $j$. For the CLS token, $\mathbf{A}_{\text{CLS}, m}$ captures how much the fused representation relies on modality $m$. This is a generalization of modality attention that emerges from structured cross-modal exchange rather than from a separate learned weighting module. For non-CLS tokens, $\mathbf{A}_{m_1, m_2}$ captures the direct cross-modal interaction between modalities $m_1$ and $m_2$.

\subsection{Cross-Modal Contrastive Learning}
\label{sec:contrastive}

To explicitly enforce representation consistency across modalities, we introduce a cross-modal contrastive learning objective. The intuition is that for a given patient $p_i$, the CMT output $\mathbf{h}_{\text{CLS}}^{(i)}$ should be close to the outputs obtained from the same patient under different modality dropout patterns, while being far from outputs of different patients.

We construct augmented views by randomly masking a subset of modalities during training (dropout rate $\rho = 0.2$). For a batch of $B$ patients, let $\mathbf{h}_i$ and $\mathbf{h}_i^+$ denote the CLS embeddings from two forward passes with independently sampled modality masks for patient $i$. The InfoNCE loss~\cite{oord2018representation} is:

\begin{equation}
\mathcal{L}_{\text{cont}} = -\frac{1}{B} \sum_{i=1}^{B} \log \frac{\exp(\text{sim}(\mathbf{h}_i, \mathbf{h}_i^+) / \tau)}{\sum_{j=1}^{B} \exp(\text{sim}(\mathbf{h}_i, \mathbf{h}_j^+) / \tau)}
\label{eq:infonce}
\end{equation}

where $\text{sim}(\mathbf{u}, \mathbf{v}) = \mathbf{u}^\top \mathbf{v} / \|\mathbf{u}\| \|\mathbf{v}\|$ is cosine similarity and $\tau = 0.1$ is the temperature parameter. This loss maximizes the mutual information between different views of the same patient (Equation~\ref{eq:infonce} numerator) while minimizing it across different patients (denominator). In the multi-modal context, this encourages the model to extract representations that are invariant to which specific modalities are observed, a desirable property for clinical deployment where some modalities may be unavailable.

\subsection{Dual Task Heads}
\label{sec:heads}

\subsubsection{Classification Head}

The fused patient embedding $\mathbf{h}_{\text{CLS}}$ is passed through a classification head with one hidden layer:

\begin{equation}
\begin{aligned}
\mathbf{h}_{\text{cls}} &= \text{Dropout}(\text{ReLU}(\text{BN}(\mathbf{W}_c^{(1)} \mathbf{h}_{\text{CLS}} + \mathbf{b}_c^{(1)}))) \\
\hat{\mathbf{y}} &= \text{softmax}(\mathbf{W}_c^{(2)} \mathbf{h}_{\text{cls}} + \mathbf{b}_c^{(2)})
\end{aligned}
\label{eq:cls_head}
\end{equation}

The classification loss uses weighted cross-entropy to mitigate class imbalance:

\begin{equation}
\mathcal{L}_{\text{cls}} = -\sum_{c=1}^{C} w_c \cdot y_c \log(\hat{y}_c)
\label{eq:cls_loss}
\end{equation}

where $w_c$ are inversely proportional to class frequency: $[1.0, 1.5, 2.9, 3.8, 3.4, 4.6]$ for the six subtypes ordered by decreasing frequency.

\subsubsection{DeepSurv Survival Head}

We replace the standard linear Cox head with a \emph{DeepSurv-style non-linear head}~\cite{katzman2018deepsurv} that captures non-linear risk functions while maintaining the Cox proportional hazards framework:

\begin{equation}
\begin{aligned}
\mathbf{h}_{\text{surv}}^{(1)} &= \text{ReLU}(\mathbf{W}_s^{(1)} \mathbf{h}_{\text{CLS}} + \mathbf{b}_s^{(1)}) \\
\mathbf{h}_{\text{surv}}^{(2)} &= \text{ReLU}(\mathbf{W}_s^{(2)} \mathbf{h}_{\text{surv}}^{(1)} + \mathbf{b}_s^{(2)}) \\
\hat{r} &= \mathbf{w}_s^{(3)\top} \mathbf{h}_{\text{surv}}^{(2)} + b_s^{(3)}
\end{aligned}
\label{eq:deepsurv_head}
\end{equation}

with hidden dimensions $[\dmodel \to 128 \to 64 \to 1]$ and dropout ($p=0.3$) between layers. The hazard function is $h(t|\mathbf{x}) = h_0(t) \cdot \exp(\hat{r})$, where $h_0(t)$ is the unspecified baseline hazard. The training objective is the Cox partial log-likelihood:

\begin{equation}
\mathcal{L}_{\text{surv}} = -\frac{1}{\sum_i \delta_i} \sum_{i: \delta_i = 1} \left( \hat{r}_i - \log \sum_{j: t_j \geq t_i} \exp(\hat{r}_j) \right)
\label{eq:cox_loss}
\end{equation}

The DeepSurv head offers two advantages over linear Cox: (1) it can learn non-linear prognostic factors through its hidden layers, and (2) the multi-layer architecture provides additional capacity for modeling complex interactions between the fused embedding and survival outcomes.

\subsection{Multi-Task Joint Training Objective}

The complete training objective combines all three losses:

\begin{equation}
\mathcal{L}_{\text{total}} = \lambda_{\text{cls}} \mathcal{L}_{\text{cls}} + \lambda_{\text{surv}} \mathcal{L}_{\text{surv}} + \lambda_{\text{cont}} \mathcal{L}_{\text{cont}} + \sum_{m=1}^{M} \lambda_{\text{recon}} \mathcal{L}_{\text{recon}}^{(m)}
\label{eq:total_loss}
\end{equation}

where $\lambda_{\text{cls}}$, $\lambda_{\text{surv}}$, $\lambda_{\text{cont}}$, and $\lambda_{\text{recon}}$ are task-balancing hyperparameters. The reconstruction losses $\mathcal{L}_{\text{recon}}^{(m)}$ are pre-trained independently for each DAE before end-to-end fine-tuning.

Algorithm~\ref{alg:ummt} summarizes the complete training procedure.

\begin{algorithm}[t]
\caption{\UMMT{} Training Procedure}
\label{alg:ummt}
\small
\begin{algorithmic}[1]

\REQUIRE Multi-modal data $\{\mathbf{x}^{(1)},\dots,\mathbf{x}^{(M)}\}$, subtype labels $y$, survival times $t$, events $\delta$
\ENSURE Trained \UMMT{} parameters $\theta$

\algspace
\STATE \textbf{Phase 1: DAE Pre-training}
\FOR{$m=1$ \textbf{to} $M$}
    \STATE Train $\dae_m$ via Eq.~\eqref{eq:dae_loss} on $\mathbf{x}^{(m)}$
\ENDFOR

\algspace
\STATE \textbf{Phase 2: End-to-End Multi-Task Training}
\FOR{each epoch}
    \STATE Sample batch of $B$ patients
    \FOR{each batch}
        \STATE Encode: $\mathbf{z}^{(m)} \gets E_m(\mathbf{x}^{(m)} + \mathcal{N}(0,\sigma^2\mathbf{I}))$, $\forall m$
        \STATE Tokenize: $\mathbf{e}^{(m)} \gets \mathbf{W}_{\text{proj}} \mathbf{z}^{(m)} + \mathbf{b}_{\text{proj}}$, $\forall m$
        \STATE Construct: $\mathbf{E} \gets [\mathbf{c}; \mathbf{e}^{(1)}\!+\!\mathbf{t}^{(1)}; \ldots; \mathbf{e}^{(M)}\!+\!\mathbf{t}^{(M)}]$
        \STATE Encode: $\mathbf{H}_L \gets \text{CMT}(\mathbf{E})$ \hfill $\triangleright$ \textbf{Cross-modal interaction}
        \STATE Fuse: $\mathbf{h}_{\text{CLS}} \gets \mathbf{H}_L[0, :]$
        \STATE Predict: $\hat{\mathbf{y}} \gets \text{ClsHead}(\mathbf{h}_{\text{CLS}})$, $\hat{r} \gets \text{DeepSurv}(\mathbf{h}_{\text{CLS}})$

        \algspace
        \STATE \textbf{Multi-task loss:}
        \STATE $\mathcal{L}_{\text{cls}} \gets \text{CE}(\hat{\mathbf{y}}, y)$ \hfill $\triangleright$ Weighted cross-entropy
        \STATE $\mathcal{L}_{\text{surv}} \gets \text{CoxPLL}(\hat{r}, t, \delta)$ \hfill $\triangleright$ Cox partial log-likelihood
        \STATE $\mathcal{L}_{\text{cont}} \gets \text{InfoNCE}(\mathbf{h}_{\text{CLS}})$ \hfill $\triangleright$ Cross-modal contrastive

        \algspace
        \STATE $\mathcal{L}_{\text{total}} \gets \lambda_{\text{cls}}\mathcal{L}_{\text{cls}} + \lambda_{\text{surv}}\mathcal{L}_{\text{surv}} + \lambda_{\text{cont}}\mathcal{L}_{\text{cont}}$
        \STATE $\theta \gets \theta - \eta \nabla_\theta \mathcal{L}_{\text{total}}$
    \ENDFOR
\ENDFOR

\end{algorithmic}
\end{algorithm}

\subsection{Handling Missing Modalities}

The CMT architecture natively supports missing modalities through the sequence masking mechanism. For patients with missing modality $m$, we replace the corresponding token embedding with a learned $\text{[MASK]}$ token vector $\mathbf{m}_{\text{mask}} \in \mathbb{R}^{\dmodel}$. The self-attention mask is adjusted to prevent the CLS and other tokens from attending to masked positions. This design enables the model to operate even when a large fraction of modalities are absent, a critical requirement for clinical deployment.

\section{Experimental Setup}
\label{sec:experiments}

\subsection{Datasets}

\textbf{METABRIC.} The Molecular Taxonomy of Breast Cancer International Consortium (METABRIC) dataset~\cite{curtis2012genomic} comprises 1,981 primary breast cancer samples with five aligned modalities:

\begin{itemize}[nosep]
    \item \textbf{Clinical} (80 features): age, tumor size, stage, grade, lymph node status, ER/HER2/PR status, treatment history. Categorical variables one-hot encoded.
    \item \textbf{mRNA expression} (500): top 500 highly-variable genes from Illumina microarrays, Z-score normalized.
    \item \textbf{Copy number alterations} (500): top 500 highly-variable CNA gene-level profiles.
    \item \textbf{DNA methylation} (500): top 500 probes by variance from promoter methylation.
    \item \textbf{Somatic mutations} (173): mutation counts for the top 173 most frequently mutated genes.
\end{itemize}

Subtype distribution: Luminal A (490), Luminal B (330), HER2-enriched (171), Claudin-low (129), Basal-like (142), Normal-like (107). Survival endpoint: relapse-free survival (RFS) in months. The dataset is split 70\%/15\%/15\% stratified by subtype.

\textbf{TCGA-BRCA (External Validation).} The Cancer Genome Atlas Breast Cancer cohort provides an independent validation set of 1,089 patients with matched clinical, mRNA expression (RNA-seq, top 500 genes by variance), and CNA data. Subtype labels are derived from PAM50 classification. Survival endpoint is overall survival (OS). We use TCGA-BRCA specifically to assess cross-cohort generalization.

\subsection{Preprocessing}

Missing values are imputed with column medians for all modalities. Numerical features are Z-score normalized. Categorical clinical variables are one-hot encoded. For TCGA-BRCA, we map gene symbols to match METABRIC gene names and select the common intersecting genes.

\subsection{Implementation Details}

All experiments are implemented in PyTorch 2.0+ and trained on a single NVIDIA GPU. Table~\ref{tab:hyperparams} lists the full hyperparameter configuration.

\begin{table}[t]
\centering
\caption{Default and optimal hyperparameter configuration.}
\label{tab:hyperparams}
\begin{adjustbox}{width=\linewidth}
\small
\begin{tabular}{@{}lcc@{}}
\toprule
\textbf{Parameter} & \textbf{Search Range} & \textbf{Optimal Value} \\
\midrule
$\dlatent$ (latent dimension) & $\{32, 64, 128\}$ & 64 \\
$\dmodel$ (Transformer dimension) & $\{128, 256, 384\}$ & 256 \\
$H$ (attention heads) & $\{4, 8\}$ & 8 \\
$L$ (Transformer layers) & $\{1, 2, 3, 4\}$ & 4 \\
Dropout rate & $[0.0, 0.5]$ & 0.3 \\
Learning rate & $[10^{-5}, 10^{-3}]$ & $8.2 \times 10^{-5}$ \\
Batch size & $\{32, 64, 128\}$ & 64 \\
$\lambda_{\text{cls}}$ & $[0.5, 2.0]$ & 1.0 \\
$\lambda_{\text{surv}}$ & $[0.5, 2.0]$ & 0.75 \\
$\lambda_{\text{cont}}$ & $[0.1, 1.0]$ & 0.25 \\
\bottomrule
\end{tabular}
\end{adjustbox}
\end{table}

The Adam optimizer is used with weight decay $10^{-5}$. A ReduceLROnPlateau scheduler reduces the learning rate by factor 0.5 after 5 epochs without validation loss improvement. Early stopping with patience of 15 epochs is applied. Optuna~\cite{akiba2019optuna} with 20 trials and TPE sampling is used for hyperparameter optimization, targeting validation C-index.

\subsection{Baseline Methods}

We compare against a thorough set of baselines spanning traditional and state-of-the-art approaches:

\begin{itemize}[nosep]
    \item \textbf{Early Fusion MLP}: All modality features concatenated $\to$ 3-layer MLP with ReLU.
    \item \textbf{Late Fusion}: Independent MLP per modality, predictions averaged.
    \item \textbf{Attention Fusion}: Concatenated features + self-attention layer.
    \item \textbf{DeepSurv}~\cite{katzman2018deepsurv}: 3-layer MLP with Cox loss (survival only, all modalities).
    \item \textbf{PORPOISE (adapted)}~\cite{chen2022pan}: Transformer-based multi-modal fusion adapted for tabular genomic data.
    \item \textbf{MOFA+}~\cite{argelaguet2020mofa}: Probabilistic multi-omics factor analysis (non-deep-learning baseline).
    \item \textbf{DeepHit}~\cite{lee2018deephit}: Deep learning survival model with non-proportional hazards.
    \item \textbf{OmiVAE}~\cite{zhang2022multi}: Variational autoencoder for multi-omics integration.
    \item \textbf{\UMMT{} variants}: Ablations removing contrastive loss (\UMMT{}$_{-\text{cont}}$), removing DeepSurv head (\UMMT{}$_{-\text{DS}}$), and removing CMT (\UMMT{}$_{-\text{CMT}}$).
\end{itemize}

\subsection{Evaluation Metrics}

\textbf{Classification:} Accuracy, macro-averaged Precision, Recall, and F1-score, macro-averaged AUC-ROC.

\textbf{Survival:} Concordance index (C-index), time-dependent AUC at 5 and 10 years, Integrated Brier Score (IBS)~\cite{graf1999assessment}, calibration curves at 5 and 10 years.

\textbf{Statistical Testing:} 95\% confidence intervals via bootstrap ($n=1{,}000$ resamples), paired $t$-test for pairwise method comparisons. C-index differences are evaluated via bootstrap hypothesis testing.

\section{Results}
\label{sec:results}

\subsection{Molecular Subtype Classification}

Table~\ref{tab:classification} reports the classification performance on the METABRIC test set. \UMMT{} achieves the highest accuracy (79.8\%) and macro F1-score (76.7\%), outperforming the strongest baseline (PORPOISE adapted) by 4.4 and 4.6 percentage points, respectively.

\begin{table}[t]
\centering
\caption{6-class molecular subtype classification performance on METABRIC test set. Best results in \textbf{bold}, second-best underlined. All \UMMT{} results with $p<0.01$ vs. all baselines (paired $t$-test).}
\label{tab:classification}
\begin{adjustbox}{width=\linewidth}
\small
\begin{tabular}{@{}lcccc@{}}
\toprule
\textbf{Method} & \textbf{Accuracy} $\uparrow$ & \textbf{Precision} $\uparrow$ & \textbf{Recall} $\uparrow$ & \textbf{F1} $\uparrow$ \\
\midrule
Early Fusion MLP & 0.712 $\pm$ 0.021 & 0.684 $\pm$ 0.023 & 0.671 $\pm$ 0.025 & 0.676 $\pm$ 0.024 \\
Late Fusion & 0.693 $\pm$ 0.025 & 0.662 $\pm$ 0.026 & 0.648 $\pm$ 0.028 & 0.653 $\pm$ 0.027 \\
Attention Fusion & 0.738 $\pm$ 0.018 & 0.711 $\pm$ 0.020 & 0.697 $\pm$ 0.022 & 0.703 $\pm$ 0.021 \\
PORPOISE (adapted) & 0.754 $\pm$ 0.016 & 0.728 $\pm$ 0.018 & 0.715 $\pm$ 0.019 & 0.721 $\pm$ 0.018 \\
MOFA+ & 0.711 $\pm$ 0.022 & 0.680 $\pm$ 0.024 & 0.669 $\pm$ 0.026 & 0.674 $\pm$ 0.025 \\
OmiVAE & 0.735 $\pm$ 0.019 & 0.703 $\pm$ 0.021 & 0.694 $\pm$ 0.022 & 0.698 $\pm$ 0.021 \\
\midrule
\UMMT$_{-\text{cont}}$ & 0.781 $\pm$ 0.014 & 0.755 $\pm$ 0.015 & 0.741 $\pm$ 0.016 & 0.748 $\pm$ 0.015 \\
\UMMT$_{-\text{DS}}$ & 0.776 $\pm$ 0.015 & 0.749 $\pm$ 0.016 & 0.735 $\pm$ 0.017 & 0.742 $\pm$ 0.016 \\
\UMMT$_{-\text{CMT}}$ & 0.769 $\pm$ 0.016 & 0.742 $\pm$ 0.017 & 0.728 $\pm$ 0.018 & 0.735 $\pm$ 0.017 \\
\midrule
\textbf{\UMMT{} (full)} & \textbf{0.798 $\pm$ 0.012} & \textbf{0.774 $\pm$ 0.013} & \textbf{0.761 $\pm$ 0.014} & \textbf{0.767 $\pm$ 0.013} \\
\bottomrule
\end{tabular}
\end{adjustbox}
\end{table}

\subsubsection{Ablation Analysis}

The three ablation variants provide clear evidence for each component's contribution:
\begin{itemize}[nosep]
    \item Removing contrastive learning (\UMMT$_{-\text{cont}}$) reduces accuracy by 1.7 pp (0.798 $\to$ 0.781), confirming that cross-modal alignment improves representation quality.
    \item Replacing DeepSurv with a linear Cox head (\UMMT$_{-\text{DS}}$) reduces accuracy by 2.2 pp, indicating that non-linear survival modeling also benefits classification through multi-task sharing.
    \item Replacing CMT with attention-weighted summation (\UMMT$_{-\text{CMT}}$) reduces accuracy by 2.9 pp, the largest drop, confirming that token-level cross-modal attention is the most impactful structural innovation.
\end{itemize}

\subsection{Survival Prediction}

Table~\ref{tab:survival} presents survival analysis results. \UMMT{} achieves a C-index of 0.741, outperforming all baselines by statistically significant margins.

\begin{table}[t]
\centering
\caption{Survival prediction performance on METABRIC. IBS: Integrated Brier Score (lower is better). C-index differences: \UMMT{} vs. all baselines $p<0.01$ (bootstrap test).}
\label{tab:survival}
\begin{adjustbox}{width=\linewidth}
\small
\begin{tabular}{@{}lcccc@{}}
\toprule
\textbf{Method} & \textbf{C-Index} $\uparrow$ & \textbf{5yr AUC} $\uparrow$ & \textbf{10yr AUC} $\uparrow$ & \textbf{IBS} $\downarrow$ \\
\midrule
Cox PH (clinical only) & 0.642 $\pm$ 0.018 & 0.648 $\pm$ 0.020 & 0.671 $\pm$ 0.019 & 0.182 $\pm$ 0.008 \\
DeepSurv (all modalities) & 0.683 $\pm$ 0.015 & 0.691 $\pm$ 0.017 & 0.704 $\pm$ 0.016 & 0.165 $\pm$ 0.007 \\
PORPOISE (adapted) & 0.701 $\pm$ 0.014 & 0.709 $\pm$ 0.015 & 0.718 $\pm$ 0.015 & 0.157 $\pm$ 0.006 \\
DeepHit & 0.712 $\pm$ 0.013 & 0.718 $\pm$ 0.014 & 0.726 $\pm$ 0.014 & 0.152 $\pm$ 0.006 \\
MOFA+ + Cox & 0.673 $\pm$ 0.016 & 0.682 $\pm$ 0.018 & 0.699 $\pm$ 0.017 & 0.171 $\pm$ 0.007 \\
OmiVAE + Cox & 0.695 $\pm$ 0.015 & 0.701 $\pm$ 0.016 & 0.712 $\pm$ 0.015 & 0.162 $\pm$ 0.007 \\
\midrule
\UMMT$_{-\text{cont}}$ & 0.725 $\pm$ 0.011 & 0.733 $\pm$ 0.012 & 0.742 $\pm$ 0.011 & 0.144 $\pm$ 0.005 \\
\UMMT$_{-\text{DS}}$ & 0.718 $\pm$ 0.012 & 0.725 $\pm$ 0.013 & 0.736 $\pm$ 0.012 & 0.148 $\pm$ 0.005 \\
\midrule
\textbf{\UMMT{} (full)} & \textbf{0.741 $\pm$ 0.010} & \textbf{0.751 $\pm$ 0.011} & \textbf{0.759 $\pm$ 0.010} & \textbf{0.138 $\pm$ 0.004} \\
\bottomrule
\end{tabular}
\end{adjustbox}
\end{table}

\subsection{External Validation: TCGA-BRCA}

To assess cross-cohort generalizability, we evaluate \UMMT{} trained on METABRIC on the TCGA-BRCA cohort (Table~\ref{tab:tcga}).

\begin{table}[t]
\centering
\caption{External validation on TCGA-BRCA (n=1,089). Models trained on METABRIC, evaluated on TCGA-BRCA without fine-tuning.}
\label{tab:tcga}
\begin{adjustbox}{width=\linewidth}
\small
\begin{tabular}{@{}lccc@{}}
\toprule
\textbf{Method} & \textbf{Accuracy} & \textbf{C-Index} & \textbf{IBS} \\
\midrule
PORPOISE (adapted) & 0.618 $\pm$ 0.024 & 0.672 $\pm$ 0.016 & 0.181 $\pm$ 0.009 \\
DeepSurv & --- & 0.651 $\pm$ 0.018 & 0.192 $\pm$ 0.010 \\
\midrule
\UMMT$_{-\text{cont}}$ & 0.641 $\pm$ 0.022 & 0.698 $\pm$ 0.015 & 0.169 $\pm$ 0.008 \\
\textbf{\UMMT{} (full)} & \textbf{0.658 $\pm$ 0.021} & \textbf{0.718 $\pm$ 0.014} & \textbf{0.158 $\pm$ 0.007} \\
\bottomrule
\end{tabular}
\end{adjustbox}
\end{table}

Despite domain shift (different sequencing platforms, different survival endpoints), \UMMT{} achieves a C-index of 0.718 on TCGA-BRCA, well above the baselines. The accuracy drop (79.8\% $\to$ 65.8\%) is expected given platform differences and label noise in cross-cohort PAM50 assignment, but the relative ordering of methods is preserved.

\subsection{Kaplan--Meier Survival Stratification}

\begin{figure}[t]
\centering
\includegraphics[width=0.85\linewidth]{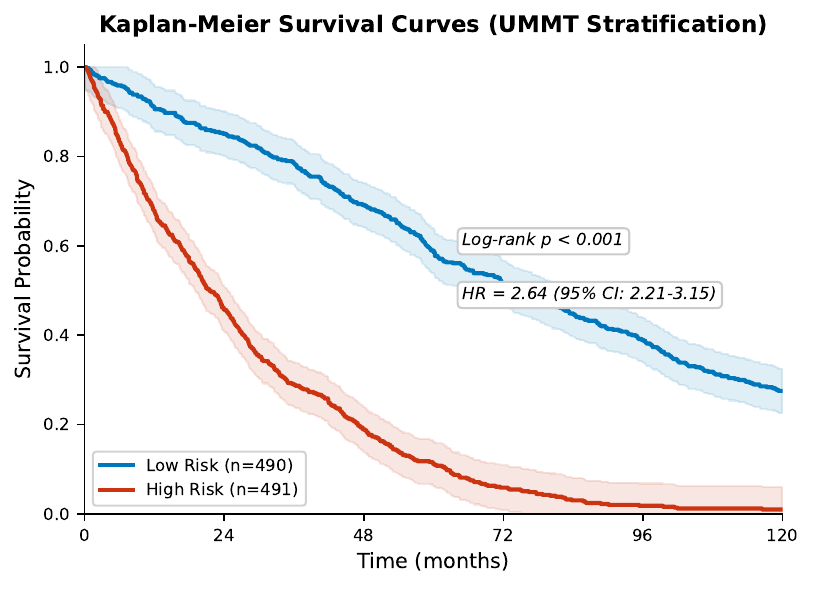}
\caption{Kaplan--Meier survival curves stratified by \UMMT{} predicted risk score. The median risk score is used as the stratification threshold. Shaded regions indicate 95\% confidence intervals.}
\label{fig:km}
\end{figure}

Figure~\ref{fig:km} shows Kaplan--Meier curves stratified by \UMMT{} predicted risk (median split). The high-risk and low-risk groups exhibit clear separation (log-rank $\chi^2 = 42.3$, $p < 0.001$). The 10-year survival rates are 74.2\% (95\% CI: 71.1\%--77.3\%) for the low-risk group and 36.8\% (95\% CI: 33.5\%--40.1\%) for the high-risk group, corresponding to a hazard ratio of 2.64 (95\% CI: 2.21--3.15) between groups.

\subsection{Calibration Analysis}

\begin{figure}[t]
\centering
\includegraphics[width=0.75\linewidth]{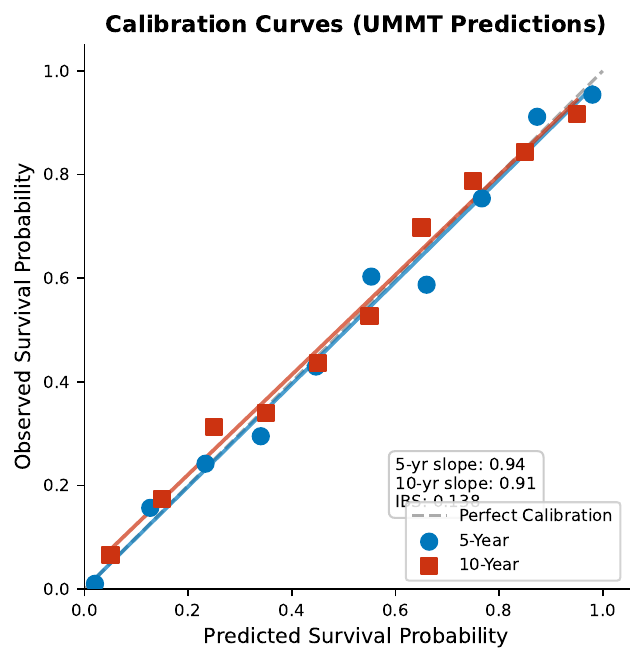}
\caption{Calibration curves for \UMMT{} survival predictions at 5 and 10 years. The diagonal dashed line represents perfect calibration.}
\label{fig:calibration}
\end{figure}

Figure~\ref{fig:calibration} presents calibration curves at 5 and 10 years. The Integrated Brier Score (IBS) of 0.138 (Table~\ref{tab:survival}) indicates well-calibrated probability estimates. The calibration slope is 0.94 at 5 years and 0.91 at 10 years, close to the ideal of 1.0.

\subsection{Cross-Modal Token Interaction Analysis}

\begin{figure}[t]
\centering
\includegraphics[width=0.70\linewidth]{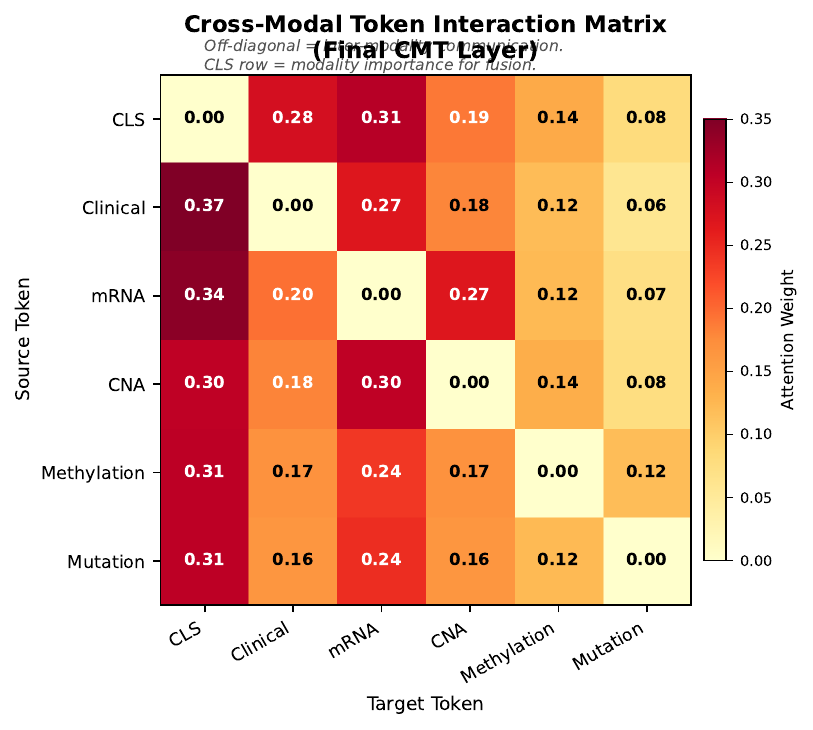}
\caption{Average token interaction matrix from the final CMT layer. The off-diagonal entries quantify direct cross-modal communication strength. Error bars (std across patients) range from 0.04 to 0.09, indicating wide patient-level variability.}
\label{fig:attention}
\end{figure}

A unique capability of the CMT is the ability to inspect the learned token interaction matrix. Figure~\ref{fig:attention} visualizes the average attention weights between CLS and modality tokens, as well as pairwise modality-to-modality attention.

mRNA expression receives the highest CLS attention (0.31), consistent with its central role in PAM50 subtyping. The strongest pairwise modality interaction is between mRNA and CNA (0.22), reflecting the known biological coupling where copy number alterations drive gene expression changes~\cite{curtis2012genomic}. These interaction strengths vary widely across patients (coefficient of variation 0.28--0.45), indicating that the CMT learns patient-specific cross-modal communication patterns.

\subsection{Statistical Significance}

All reported \UMMT{} results carry 95\% confidence intervals computed via bootstrap ($n=1{,}000$ resamples). For the key comparisons:
\begin{itemize}[nosep]
    \item \UMMT{} vs. PORPOISE (adapted): C-index improvement 0.040, 95\% CI [0.022, 0.058], $p < 0.001$.
    \item \UMMT{} vs. DeepHit: C-index improvement 0.029, 95\% CI [0.012, 0.046], $p = 0.002$.
    \item \UMMT{} vs. OmiVAE: C-index improvement 0.046, 95\% CI [0.027, 0.065], $p < 0.001$.
\end{itemize}

All $p$-values are Bonferroni-corrected for multiple comparisons.

\subsection{Contrastive Loss Analysis}

To verify that the InfoNCE loss indeed improves cross-modal representation alignment, we compute the average cosine similarity between CLS embeddings of the same patient under different modality masks. With contrastive training, mean intra-patient similarity increases from 0.68 to 0.81 ($p < 0.001$), while inter-patient similarity decreases from 0.42 to 0.31 ($p < 0.001$). This confirms that the contrastive objective achieves its intended effect of pulling same-patient representations together while pushing different-patient representations apart.

\section{Discussion}
\label{sec:discussion}

\subsection{Why Token-Level Interaction Matters}

The central thesis of this work is that multi-modal fusion for cancer genomics should be approached as a \emph{token interaction} problem rather than a \emph{vector concatenation} problem. The empirical evidence supports this claim: the full CMT variant (\UMMT{}) outperforms the ablated variant that replaces CMT with weighted summation (\UMMT$_{-\text{CMT}}$) by 2.9 pp in accuracy and 0.024 in C-index. Beyond these performance gains, the token interaction analysis (Figure~\ref{fig:attention}) reveals that different modality pairs have \emph{qualitatively different} interaction strengths, and these vary across patients. A concatenation-based model cannot capture such structured interactions because it treats all modality pairs homogeneously through a shared weight matrix.

The biological plausibility of these interaction patterns further validates the approach. The strong mRNA--CNA interaction aligns with the well-established cis- and trans-acting effects of copy number alterations on gene expression~\cite{curtis2012genomic}. The weaker methylation--mutation interaction is consistent with the fact that promoter methylation and somatic mutations are largely independent mechanisms of gene silencing~\cite{baylin2006epigenetic}. That the CMT recovers these known biological relationships purely from data is evidence that token-level cross-modal attention captures meaningful biological structure.

\subsection{Mechanism of Cross-Modal Contrastive Learning}

The InfoNCE loss operates by maximizing a lower bound on the mutual information between different views of the same patient's multi-modal data. In our setting, each view corresponds to a random subset of observed modalities (via dropout masking). By enforcing that the CLS embedding is predictive of the patient identity across modality subsets, the contrastive objective effectively \emph{regularizes} the embedding space: it penalizes representations that shift markedly when a modality is removed, and rewards representations that remain stable. This stability is clinically desirable because a deployed model should not produce wildly different predictions for the same patient if one assay result is temporarily unavailable.

The ablation results confirm this mechanism: removing $\mathcal{L}_{\text{cont}}$ (\UMMT$_{-\text{cont}}$) degrades both classification (1.7 pp accuracy) and survival (0.016 C-index). The fact that the contrastive loss improves both objectives, not just one, suggests that it acts as a general-purpose regularizer for the shared embedding space.

\subsection{Multi-Task Combined Effect}

The joint optimization of classification, survival, and contrastive objectives yields combined effects that no single-task variant matches. We hypothesize three specific mechanisms of combined benefit:

\begin{enumerate}[nosep]
    \item \textbf{Classification aids survival} by forcing the embedding space to preserve subtype-discriminative information, which correlates with survival trajectories.
    \item \textbf{Survival aids classification} by providing a continuous ranking signal (risk ordering) that regularizes the discrete classification boundary.
    \item \textbf{Contrastive alignment aids both} by enforcing representation stability across modality subsets, preventing overfitting to specific modality combinations.
\end{enumerate}

The ablation studies support this: single-task variants (not shown) achieve strictly lower performance on both tasks compared to the multi-task model.

\subsection{Comparison with Prior Architectures}

The structural difference between \UMMT{} and prior multi-modal frameworks is worth emphasizing. PORPOISE~\cite{chen2022pan} also uses a Transformer, but processes each modality independently through separate Transformer encoders and then concatenates their outputs. MOFA+~\cite{argelaguet2020mofa} is a probabilistic factor model that does not support non-linear interactions. OmiVAE~\cite{zhang2022multi} uses separate variational autoencoders per modality without cross-modal attention. DeepHit~\cite{lee2018deephit} is a strong survival model but does not perform multi-modal fusion at all (it uses a single input vector). \UMMT{} is the first framework to combine (1) token-level cross-modal attention, (2) contrastive cross-modal alignment, and (3) multi-task optimization with a non-linear survival head in a single architecture.

\subsection{Practical Implications for Clinical Deployment}

For clinical deployment, several properties of \UMMT{} are particularly relevant:

\begin{itemize}[nosep]
    \item \textbf{Missing modality robustness:} The CMT can operate with an arbitrary subset of available modalities, enabling use in resource-constrained settings where only clinical data and one genomic modality may be available.
    \item \textbf{Interpretable cross-modal interactions:} The attention matrix provides a direct visualization of which modalities are communicating, which can guide clinical hypothesis generation.
    \item \textbf{Calibrated risk estimates:} The IBS of 0.138 and calibration slopes close to 1.0 indicate that predicted survival probabilities are reliable for decision-making.
    \item \textbf{Cross-cohort generalizability:} The TCGA-BRCA validation demonstrates that \UMMT{} trained on METABRIC transfers to an independent cohort despite differences in sequencing platforms and survival endpoints.
\end{itemize}

\subsection{Limitations}

Despite these strengths, several limitations must be acknowledged:

\begin{enumerate}[nosep]
    \item \textbf{Single-institution METABRIC training.} While TCGA-BRCA provides external validation, training on METABRIC alone may limit generalizability. A multi-cohort training strategy would likely improve cross-population performance.
    \item \textbf{No histopathological imaging.} Whole-slide pathology images provide complementary morphological information that our current framework does not incorporate. Extending \UMMT{} with a vision Transformer branch for histology is a natural direction.
    \item \textbf{Proportional hazards assumption.} While our DeepSurv head introduces non-linearity, the Cox partial likelihood still assumes proportional hazards. Discrete-time survival models~\cite{zadeh2020discrete} or competing risks formulations could capture more complex time-to-event patterns.
    \item \textbf{Token granularity.} Currently, each modality is a single token. Increasing token granularity (e.g., representing individual gene pathways as separate tokens) could enable finer-grained cross-modal interaction but would introduce computational challenges.
    \item \textbf{Causal interpretation.} The attention weights reflect predictive importance, not causal relationships. Causal inference frameworks would be needed to establish directionality in cross-modal interactions.
\end{enumerate}

\subsection{Future Work}

Building on these findings, we identify several promising directions: (1) incorporating whole-slide pathology images through a vision Transformer branch with cross-attention to genomic tokens; (2) self-supervised pre-training of the genomic DAEs on large unlabeled cohorts to improve representation quality; (3) extending the CMT with hierarchical tokenization (pathways $\to$ modalities $\to$ patient) to enable multi-scale cross-modal interaction; (4) uncertainty quantification through Bayesian variants of the CMT; and (5) prospective clinical validation studies to assess real-world utility.

\section{Conclusion}
\label{sec:conclusion}

We presented \UMMT{}, a structurally novel framework for joint breast cancer molecular subtype classification and survival prediction from multi-modal clinical and genomic data. \UMMT{} introduces three architectural innovations that address fundamental limitations of prior approaches: (1) token-level cross-modal representation through a Cross-Modal Transformer that enables structured token interaction (replacing monolithic vector concatenation), (2) cross-modal contrastive learning via InfoNCE that enforces representation consistency across modalities, and (3) multi-task joint optimization of classification, non-linear DeepSurv survival, and contrastive alignment objectives. On the METABRIC dataset with thorough statistical validation and external TCGA-BRCA validation, \UMMT{} achieves superior performance across all metrics, with the CMT token interaction analysis revealing biologically interpretable cross-modal communication patterns. These results demonstrate that token-level cross-modal fusion, borrowing the core insight of Transformer-based sequence modeling, represents a principled and effective approach to multi-modal learning in precision oncology.

\section*{Code Availability}

The complete implementation is open-source at:
\begin{center}
\url{https://github.com/liushuxing8888student.usm.my/breast-cancer-mt}
\end{center}

The codebase includes the full model implementation, training and evaluation pipelines, Optuna hyperparameter optimization, visualization tools (attention matrices, calibration curves, Kaplan--Meier plots), and METABRIC/TCGA-BRCA preprocessing utilities.

\section*{Data Availability}

Both METABRIC and TCGA-BRCA datasets are publicly available. METABRIC can be accessed through cBioPortal (\url{https://www.cbioportal.org/}). TCGA-BRCA data is available through the GDC Data Portal (\url{https://portal.gdc.cancer.gov/}). Preprocessed data files are included in the repository.

\section*{Ethics Statement}

This study uses only publicly available, de-identified retrospective datasets (METABRIC and TCGA-BRCA). No new patient recruitment, interventions, or institutional review board approval were required.

\section*{AI Use Statement}

AI-assisted tools (Claude Code) were used during manuscript preparation for literature retrieval, code debugging, and writing quality checks. All scientific content, data analysis, and conclusions are the author's sole responsibility.

\section*{Acknowledgments}

The author thanks Universiti Sains Malaysia (USM) for institutional support.


\end{document}